\documentclass[10pt,twocolumn,letterpaper]{article}

\usepackage[pagenumbers]{cvpr} % To force page numbers, e.g. for an arXiv version

\usepackage{graphicx}
\usepackage{amsmath}
\usepackage{amssymb}
\usepackage{pifont} % for cmark
\usepackage{enumitem}
\usepackage{booktabs}

\usepackage[dvipsnames,svgnames,x11names]{xcolor}

\definecolor{citecolor}{HTML}{0071bc}
\usepackage[pagebackref=true,breaklinks=true,colorlinks,citecolor=citecolor,bookmarks=false]{hyperref}

\usepackage{textpos}  % for superposing a text box

\usepackage[capitalize]{cleveref}
\crefname{section}{Sec.}{Secs.}
\Crefname{section}{Section}{Sections}
\Crefname{table}{Table}{Tables}
\crefname{table}{Tab.}{Tabs.}

\newcommand{\cmark}{\ding{51}}

\newlength\savewidth\newcommand\shline{\noalign{\global\savewidth\arrayrulewidth
  \global\arrayrulewidth 1pt}\hline\noalign{\global\arrayrulewidth\savewidth}}
\newcommand{\tablestyle}[2]{\setlength{\tabcolsep}{#1}\renewcommand{\arraystretch}{#2}\centering\small}
\makeatletter\renewcommand\paragraph{\@startsection{paragraph}{4}{\z@}
  {.5em \@plus1ex \@minus.2ex}{-.5em}{\normalfont\normalsize\bfseries}}\makeatother
\newcommand{\customfootnotetext}[2]{{% Group to localize change to footnote
  \renewcommand{\thefootnote}{#1}% Update footnote counter representation
  \footnotetext[0]{#2}}}% Print footnote text

\def\cvprPaperID{****} % *** Enter the CVPR Paper ID here
\def\confName{CVPR}
\def\confYear{2023}

\newcommand{\ourmethod}{ODISE}

\begin{document}

\title{Open-Vocabulary Panoptic Segmentation with Text-to-Image Diffusion Models}

\author{
Jiarui Xu\textsuperscript{1*} \hspace{2mm}
Sifei Liu\textsuperscript{2$\dagger$} \hspace{2mm}
Arash Vahdat\textsuperscript{2$\dagger$} \hspace{2mm}
Wonmin Byeon\textsuperscript{2}  
\\
Xiaolong Wang\textsuperscript{1} \hspace{2mm}
Shalini De Mello\textsuperscript{2} \\
\vspace{1mm}
\textsuperscript{1}UC San Diego \qquad \textsuperscript{2}NVIDIA \\ 
}

\twocolumn[{%
\renewcommand\twocolumn[1][]{#1}%
\maketitle
\begin{center}
    \centering
    \captionsetup{type=figure}
    \vspace{-10pt}
    \includegraphics[width=\textwidth]{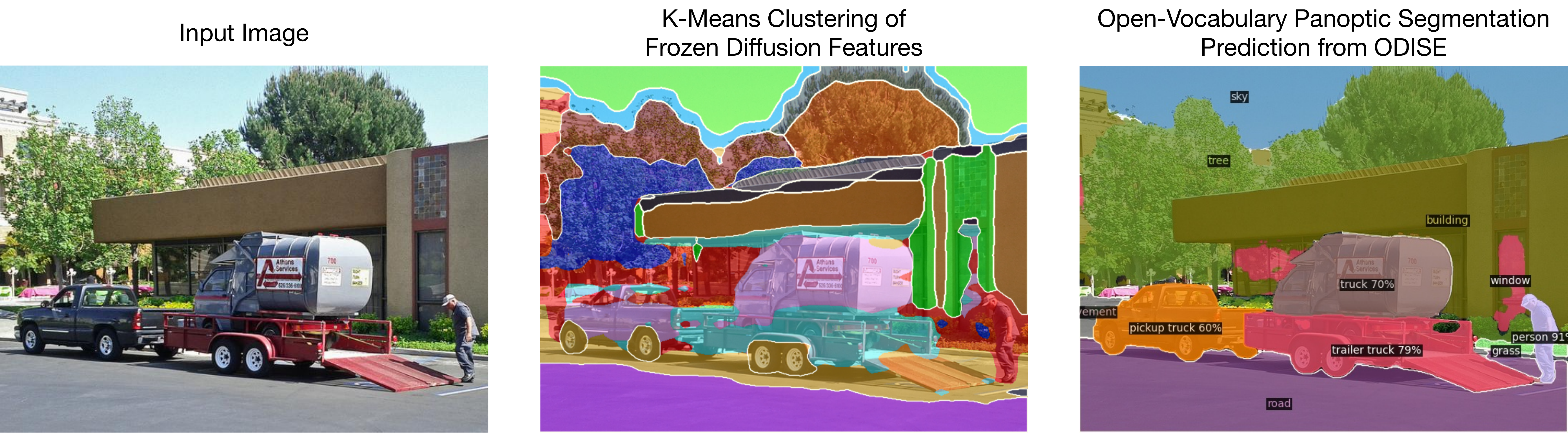}
    \vspace{-18pt}
    \caption{
    \label{fig:teaser}
    We learn open-vocabulary panoptic segmentation with the internal representation of text-to-image diffusion models. K-Means clustering of the diffusion model's internal representation shows semantically differentiated and localized information wherein objects are well grouped together (middle figure). We leverage these dense and rich diffusion features to perform open-vocabulary panoptic segmentation (right figure).}
\end{center}%
}]
\customfootnotetext{*}{Jiarui Xu was an intern at NVIDIA during the project. ${\dagger}$ equal contribution. }

\vspace{-.5em}
\begin{abstract}
\vspace{-.5em}

We present \ourmethod{}: Open-vocabulary DIffusion-based panoptic SEgmentation, which unifies pre-trained text-image diffusion and discriminative models to perform open-vocabulary panoptic segmentation. 
Text-to-image diffusion models have the remarkable ability to generate high-quality images with diverse open-vocabulary language descriptions. 
This demonstrates that their internal representation space is highly correlated with open concepts in the real world. 
Text-image discriminative models like CLIP, on the other hand, are good at classifying images into open-vocabulary labels. 
We leverage the frozen internal representations of both these models to perform panoptic segmentation of any category in the wild. 
Our approach outperforms the previous state of the art by significant margins on both open-vocabulary panoptic and semantic segmentation tasks. 
In particular, with COCO training only, our method achieves 23.4 PQ and 30.0 mIoU on the ADE20K dataset, with 8.3 PQ and 7.9 mIoU absolute improvement over the previous state of the art. 
We open-source our code and models at \url{https://github.com/NVlabs/ODISE}.
\end{abstract}
\vspace{-1.5em}

% hack: add link to project page
\begin{textblock*}{.9\textwidth}[.5,0](0.5\textwidth, -.967\textwidth)
    \centering
    {Project Page: \url{https://jerryxu.net/ODISE/}}
\end{textblock*}

\section{Introduction}

Humans look at the world and can recognize limitless categories. Given the scene presented in Fig.~\ref{fig:teaser}, besides identifying every vehicle as a ``truck'', we immediately understand that one of them is a pickup truck requiring a trailer to move another truck. To reproduce an intelligence with such a fine-grained and unbounded understanding, the problem of open-vocabulary recognition~\cite{radford2021clip,li2022language,zhou2022detic,xu2022groupvit} has recently attracted a lot of attention in computer vision. However, very few works are able to provide a unified framework that parses all object instances and scene semantics at the same time, i.e., panoptic segmentation. 

Most current approaches for open-vocabulary recognition rely on the excellent generalization ability of text-image discriminative models~\cite{radford2021clip, jia2021align} trained with Internet-scale data. While such pre-trained models are good at classifying individual object proposals or pixels, they are not necessarily optimal for performing scene-level structural understanding. Indeed, it has been shown that CLIP~\cite{radford2021clip} often confuses the spatial relations between objects~\cite{subramanian2022reclip}. We hypothesize that the lack of spatial and relational understanding in text-image discriminative models is a bottleneck for open-vocabulary panoptic segmentation.

On the other hand, text-to-image generation using diffusion models trained on Internet-scale data~\cite{rombach2022ldm,saharia2022imagen,ramesh2022dalle2, balaji2022eDiffI, zhou2022latte} has recently revolutionized the field of image synthesis. It offers unprecedented image quality, generalizability, composition-ability and, semantic control via the input text. An interesting observation is that to condition the image generation process on the provided text, diffusion models compute cross-attention between the text's embedding and their internal visual representation. This design implies the plausibility of the internal representation of diffusion models being well-differentiated and correlated to high/mid-level semantic concepts that can be described by language. 
As a proof-of-concept, in Fig.\ref{fig:teaser} (center), we visualize the results of clustering a diffusion model's internal features for the image on the left. While not perfect, the discovered groups are indeed semantically distinct and localized. 
Motivated by this finding, we ask the question of whether Internet-scale text-to-image diffusion models can be exploited to create universal open-vocabulary panoptic segmentation learner for any concept in the wild?

To this end, we propose \textit{\ourmethod{}}: Open-vocabulary DIffusion-based panoptic SEgmentation (pronounced \emph{o-di-see}), a model that leverages both large-scale text-image diffusion and discriminative models to perform state-of-the-art panoptic segmentation of any category in the wild. An overview of our approach is illustrated in Fig.~\ref{fig:pipeline}. At a high-level it contains a pre-trained frozen text-to-image diffusion model into which we input an image and its caption and extract the diffusion model's internal features for them. With these features as input, our mask generator produces panoptic masks of all possible concepts in the image. We train the mask generator with annotated masks available from a training set. A mask classification module then categorizes each mask into one of many open-vocabulary categories by associating each predicted mask's diffusion features with text embeddings of several object category names. We train this classification module with either mask category labels or image-level captions from the training dataset.
Once trained, we perform open-vocabulary panoptic inference with both the text-image diffusion and discriminative models to classify a predicted mask. On many different benchmark datasets and across several open-vocabulary recognition tasks, \ourmethod{} achieves state-of-the-art accuracy outperforming the existing baselines by large margins.

Our contributions are the following:
\begin{itemize}[noitemsep,nosep]
\item To the best of our knowledge, \ourmethod{} is the first work to explore large-scale text-to-image diffusion models for open-vocabulary segmentation tasks.
\item We propose a novel pipeline to effectively leverage both text-image diffusion and discriminative models to perform open-vocabulary panoptic segmentation. 
\item We significantly advance the field forward by outperforming all existing baselines on many open-vocabulary recognition tasks, and thus establish a new state of the art in this space.
\end{itemize}

\section{Related Work}

\textbf{Panoptic Segmentation.}
Panoptic segmentation~\cite{kirillov2019panoptic} is a fundamental vision task that encompasses both instance and semantic segmentation. However, previous works~\cite{cheng2022mask2former, cheng2020panopticdeeplab, kirillov2019panoptic, wang2021maxdeeplab, carion2020detr,li2022maskdino, yu2022kmeansmaskformer,li2022panopticsegformer,zhang2021knet, cheng2021maskformer, ren2021refine,li2021panopticfcn} follow a closed closed-vocabulary assumption and only recognize categories present in the training set. They are hence limited in segmenting things/stuff present in finite-sized vocabularies, which are much smaller than the typical vocabularies that we use to describe the real world. 

\textbf{Open-Vocabulary Segmentation.} 
Most prior works on open-vocabulary segmentation either perform object detection with instance segmentation alone~\cite{gu2021vild,zhou2022detic, du2022detpro, ghiasi2021open, minderer2022simple, zhong2022regionclip,zareian2021ovrccn,li2022glip,zang2022ovdetr} or open-vocabulary semantic segmentation alone~\cite{li2022language,ghiasi2021open, xu2022groupvit,zhou2021denseclip}. In contrast, we propose a novel unified framework for both open-vocabulary instance and semantic segmentation. Another distinction is that prior works only use large-scale models pre-trained for image discriminative tasks, e.g., image classification~\cite{he2016resnet, liu2021swin} or image-text contrastive learning~\cite{radford2021clip, jia2021align, mu2021slip, li2021supervision}. The concurrent work MaskCLIP\cite{ding2022open} also uses CLIP~\cite{radford2021clip}. However, such discriminative models' internal representations are sub-optimal for performing segmentation tasks versus those derived from image-to-text diffusion models as shown in our experiments. %We, instead, directly leverage the internal representations of frozen diffusion-based text-to-image generative models. 

\textbf{Generative Models for Segmentation.} There exist prior works, which are similar in spirit to ours in their use of image generative models, including GANs~\cite{karras2019stylegan,brock2018biggan,esser2021vqgan, karras2020styleganv2, zhu2017cyclegan} or diffusion models~\cite{ho2020denoising, song2020score, song2020ddim, sohl2015deep, song2019generative, vahdat2021score, dockhorn2021score, jolicoeur2020adversarial, song2020improved, dhariwal2021diffusion, nichol2021improved} to perform semantic segmentation~\cite{zhang2021datasetgan,li2022bigdatasetgan,baranchuk2021ddpmseg, tritrong2021repurposing, galeev2021learning, meng2021sdedit}. They first train generative models on small-vocabulary datasets, e.g.,  cats~\cite{yu2015lsun}, human faces~\cite{karras2019stylegan} or ImageNet~\cite{deng2009imagenet} and then with the help of few-shot hand-annotated examples per category, learn to classify the internal representations of the generative models into semantic regions. They either synthesize many images and their mask labels to train a separate segmentation network~\cite{zhang2021datasetgan,li2022bigdatasetgan}; or directly use the generative model to perform segmentation~\cite{baranchuk2021ddpmseg}. Among them, DDPMSeg~\cite{baranchuk2021ddpmseg} shows the state-of-the-art accuracy. These prior works introduce the key idea that the internal representations of generative models may be sufficiently differentiated and correlated to mid/high-level visual semantic concepts and could be used for semantic segmentation. Our work is inspired by them, but it is also different in many respects. While previous works primarily focus on label-efficient semantic segmentation of small closed vocabularies, we, on the other hand, tackle open-vocabulary panoptic segmentation of many more and unseen categories in the wild.

\section{Method}

\begin{figure*}[t]
    \centering
    \includegraphics[width=\linewidth]{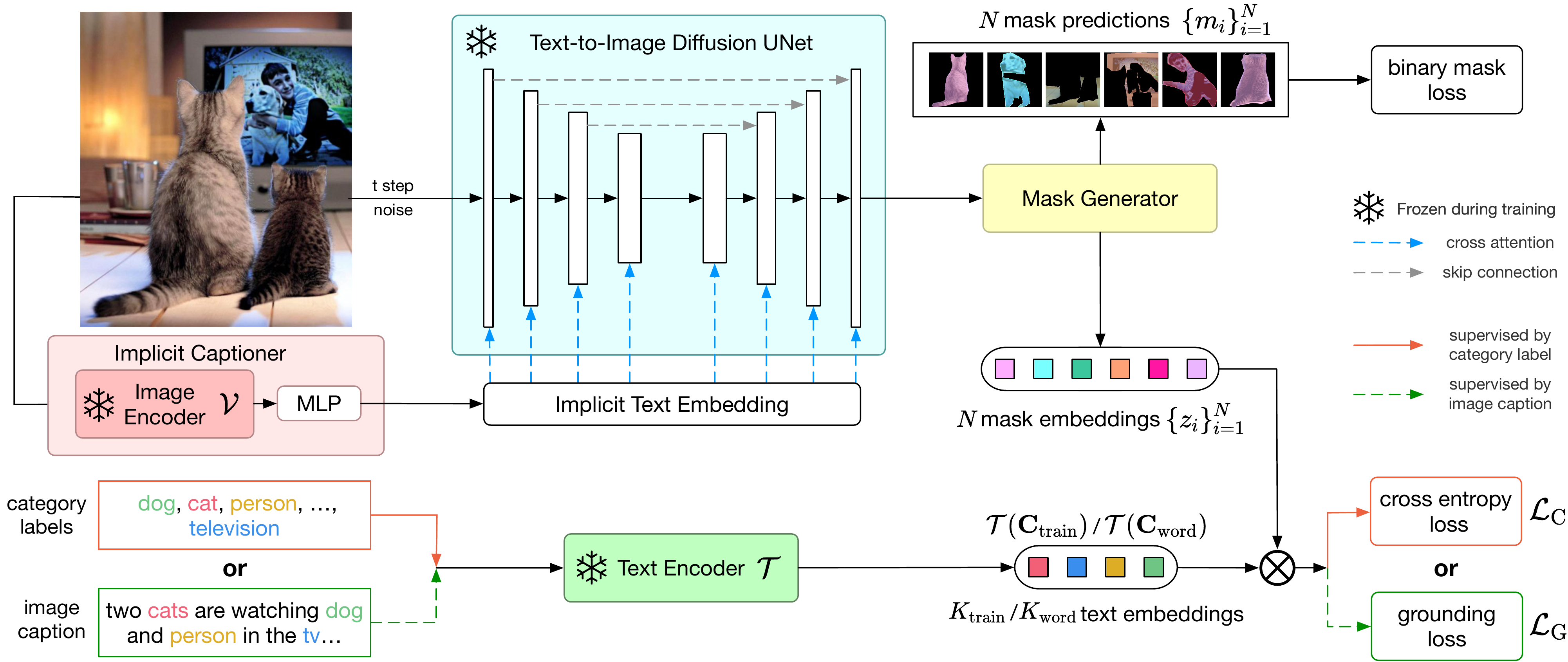}
    \vspace{-1.5em}
    \caption{
        \textbf{\ourmethod{} Overview and Training Pipeline}. 
        We first encode the input image into an implicit text embedding with an implicit captioner (image encoder $\mathcal{V}$ and MLP). 
        With the image and its implicit text embedding as input, we extract their diffusion features from a frozen text-to-image diffusion UNet (Sec~\ref{sec:diffusion}). 
        With the UNet's features, a mask generator predicts class-agnostic binary masks and their associated mask embedding features (Sec~\ref{sec:mask-generator}).
        We perform a dot product between the mask embedding features and the text embeddings of training category names (\textcolor[HTML]{EC603D}{red box}) or the nouns of the image's caption (\textcolor[HTML]{008F00}{green box}) to categorize them.
        The similarity matrix for mask classification is supervised by either a cross entropy loss with ground truth category labels (\textcolor[HTML]{EC603D}{red solid path}), or via a grounding loss with the paired image captions (\textcolor[HTML]{008F00}{green dash path}) (Sec~\ref{sec:discriminative}).
    }
    \vspace{-1.5em}
    \label{fig:pipeline}
\end{figure*}
\vspace{-0.25em}
\subsection{Problem Definition}
\vspace{-0.25em}
\label{sec:preliminaries}
Following~\cite{kirillov2019panoptic, ding2022open}, we train a model with a set of base training categories $\mathbf{C}_{\text{train}}$, which may be different from the test categories, $\mathbf{C}_{\text{test}}$, i.e., $\mathbf{C}_{\text{train}} \neq \mathbf{C}_{\text{test}}$. $\mathbf{C}_{\text{test}}$ may contain novel categories not seen during training. We assume that during training, the binary panoptic mask annotation for each category in an image is provided. Additionally, we also assume that either the category label of each mask or a text caption for the image is available. During testing, neither the category label nor the caption is available for any image, and only the names of the test categories $\mathbf{C}_{\text{test}}$ are provided.

\subsection{Method Overview}
\begin{figure*}[t]
    \centering
    \includegraphics[width=\linewidth]{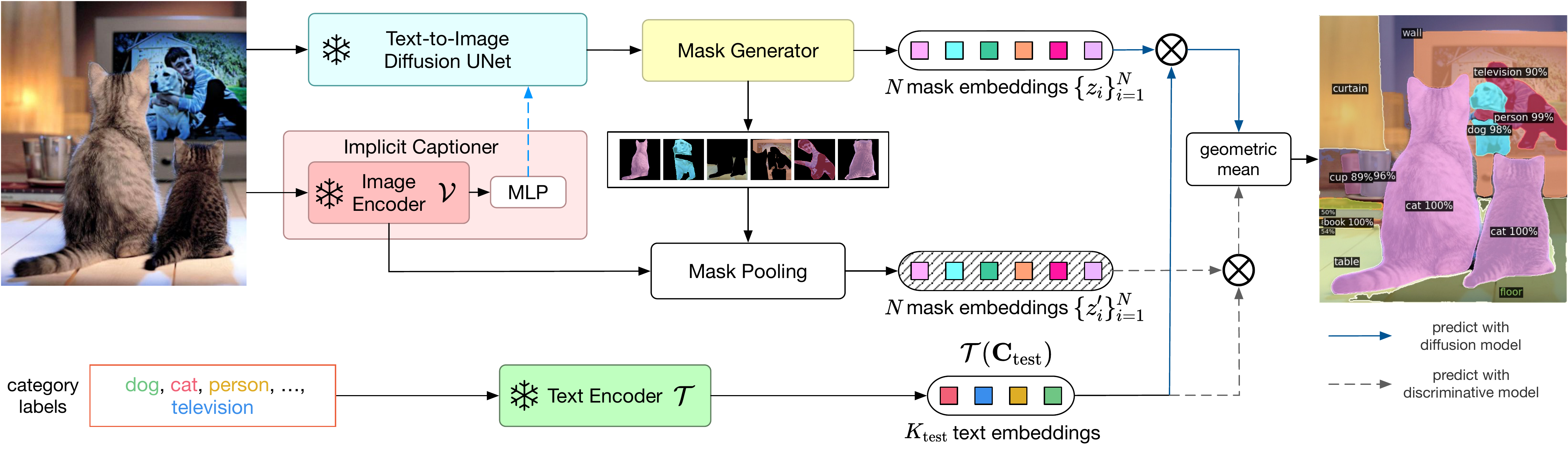}
    \vspace{-1.5em}
    \caption{
        \textbf{Open-Vocabulary Inference Pipeline}.  
        To classify each mask embedding into testing categories $\mathbf{C}_\text{test}$, we compute its similarity with the text encoder $\mathcal{T}$ embedding of category names.
        Besides the mask embeddings from text-to-image diffusion model $\{z_i\}_{i=1}^N$, we also perform mask pooling on the features of image encoder $\mathcal{V}$ from text-image discriminative model to get $\{z'_i\}_{i=1}^N$.
        We fuse the prediction of diffusion model (\textcolor[HTML]{005493}{blue solid path}) and discriminative model (\textcolor[HTML]{7E7E7E}{grey dash path}) with geometric mean. 
    }
    \vspace{-.5em}
    \label{fig:inference}
\end{figure*}

\vspace{-0.25em}
An overview of our method \ourmethod{}, for open-vocabulary panoptic segmentation of any category in the wild is shown in Fig.~\ref{fig:pipeline}. At a high-level, it contains a text-to-image diffusion model into which we input an image and its caption and extract the diffusion model's internal features for them (Sec~\ref{sec:diffusion}).
With these extracted features as input, and the provided training mask annotations, we train a mask generator to generate panoptic masks of all possible categories in the image (Sec~\ref{sec:mask-generator}). Using the provided training images' category labels or text captions, we also train an open-vocabulary mask classification module. It uses each predicted mask's diffusion features along with a text encoder's embeddings of the training category names to classify a mask (Sec~\ref{sec:discriminative}). 
Once trained, we perform open-vocabulary panoptic inference with both the text-image diffusion and discriminative models (Sec~\ref{sec:inference} and Fig.~\ref{fig:inference}). In the following sections, we describe each of these components.

\subsection{Text-to-Image Diffusion Model}
\label{sec:diffusion}
\vspace{-0.25em}
We first provide a brief overview of text-to-image diffusion models and then describe how we extract features from them for panoptic segmentation. 

\paragraph{Background}

A text-to-image diffusion model can generate high-quality images from provided input text prompts. 
It is trained with millions of image-text pairs crawled from the Internet~\cite{nichol2021glide, ramesh2022dalle2, saharia2022imagen}. 
The text is encoded into a text embedding with a pre-trained text encoder, \textit{e.g.}, T5~\cite{raffel2020t5} or CLIP~\cite{radford2021clip}.
Before being input into the diffusion network, an image is distorted by adding some level of Gaussian noise to it. 
The diffusion network is trained to undo the distortion given the noisy input and its paired text embedding.
During inference, the model takes image-shaped pure Gaussian noise and the text embedding of a user provided description as input, and progressively de-noises it to a realistic image via several iterations of inference. 

\paragraph{Visual Representation Extraction}
The prevalent diffusion-based text-to-image generative models~\cite{ramesh2022dalle2,saharia2022imagen,rombach2022ldm, nichol2021glide} typically use a UNet architecture to learn the denoising process. As shown in the blue block in Fig.~\ref{fig:pipeline}, the UNet consists of convolution blocks, upsampling and downsampling blocks, skip connections and attention blocks, which perform cross-attention~\cite{vaswani2017attention} between a text embedding and UNet features.
At every step of the de-noising process, diffusion models use the text input to infer the de-noising direction of the noisy input image. 
Since the text is injected into the model via cross attention layers, it encourages visual features to be correlated to rich semantically meaningful text descriptions. 
Thus the feature maps output by the UNet blocks can be regarded as rich and dense features for panoptic segmentation. 

Our method only requires a single forward pass of an input image through the diffusion model to extract its visual representation, as opposed to going through the entire multi-step generative diffusion process.  
Formally, given an input image-text pair $(x, s)$, 
we first sample a noisy image $x_t$ at time step $t$ as:
\begin{equation}
\label{eqn:noise}
x_t \triangleq \sqrt{\bar{\alpha}_{t}} x+\sqrt{1-\bar{\alpha}_{t}} \epsilon, \quad \epsilon \sim \mathcal{N}(0,\mathbf{I}),
\end{equation}
where $t$ is the diffusion step we use, $\alpha_1,\dots,\alpha_T$ represent a pre-defined noise schedule where $\bar{\alpha}_{t}{=}\prod_{k=1}^t\alpha_k$, as defined in~\cite{ho2020denoising}.
We encode the caption $s$ with a pre-trained text encoder $\mathcal{T}$ and extract the text-to-image diffusion UNet's internal features $f$ for the pair by feeding it into the UNet
\begin{equation}
   f =  \text{UNet}(x_t, \mathcal{T}(s)).
\end{equation}

It is worth noting that the diffusion model's visual representation $f$ for $x$ is dependent on its paired caption $s$. 
It can be extracted correctly when paired image-text data is available, e.g., during pre-training of the text-to-image diffusion model. 
However, it becomes problematic when we want to extract the visual representation of images without paired captions available, which is the common use case for our application. 
For an image without a caption, we could use an empty text as its caption input, but that is clearly suboptimal, which we also show in our experiments. 
In what follows, we introduce a novel \emph{Implicit Captioner} that we design to overcome the need for explicitly captioned image data. It also yields optimal downstream task performance.

\paragraph{Implicit Captioner}
Instead of using an off-the-shelf captioning network to generate captions, we train a network to generate an implicit text embedding from the input image itself. We then input this text embedding into the diffusion model directly.
We name this module an implicit captioner.
The red block in Fig.~\ref{fig:pipeline} shows the architecture of the implicit captioner.
Specifically, to derive the implicit text embedding for an image, we leverage a pre-trained frozen image encoder $\mathcal{V}$, e.g., from CLIP~\cite{radford2021clip} to encode the input image $x$ into its embedding space. 
We further use a learned MLP to project the image embedding into an implicit text embedding, which we input into text-to-image diffusion UNet. During open-vocabulary panoptic segmentation training, the parameters of the image encoder and of the UNet are unchanged and we only fine-tune the parameters of the MLP.

Finally, the text-to-image diffusion model's UNet along with the implicit captioner, together form \ourmethod{}'s feature extractor that computes the visual representation $f$ for an input image $x$. Formally, we compute the visual representation $f$ as:
\begin{equation}
    \begin{split}
    f &=  \text{UNet}(x_t, \text{ImplicitCaptioner}(x))\\
        &=  \text{UNet}(x_t, \text{MLP}\circ \mathcal{V}(x)).
    \end{split}
\end{equation}

\subsection{Mask Generator}
\label{sec:mask-generator}
The mask generator takes the visual representation $f$ as input and outputs $N$ class-agnostic binary masks $\{m_i\}_{i=1}^N$ and their corresponding $N$ mask embedding features $\{z_i\}_{i=1}^N$.
The architecture of the mask generator is not restricted to a specific one. 
It can be any panoptic segmentation network capable of generating mask predictions of the whole image. 
We can instantiate our method with both bounding box-based~\cite{carion2020detr, kirillov2019panopticfpn} and direct segmentation mask-based~\cite{cheng2020panopticdeeplab, cheng2021maskformer, cheng2022mask2former, wang2021maxdeeplab} methods.
While using bounding box-based methods like ~\cite{carion2020detr, kirillov2019panopticfpn}, we can pool the ROI-Aligned~\cite{he2017maskrcnn} features of each predicted mask's region to compute its mask embedding features.
For segmentation mask-based methods like ~\cite{cheng2020panopticdeeplab, cheng2021maskformer, cheng2022mask2former, wang2021maxdeeplab}, we can directly perform masked pooling on the final feature maps to compute the mask embedding features. 
Since our representation focuses on dense pixel-wise predictions, we use a direct segmentation-based architecture. 
Following~\cite{he2017maskrcnn}, we supervise the predicted class-agnostic binary masks via a pixel-wise binary cross entropy loss along with their corresponding ground truth masks (treated as class-agnostic ones as well).
Next, we describe how we classify each mask, represented by its mask embedding feature, into an open vocabulary.  

\subsection{Mask Classification}
\label{sec:discriminative}

To assign each predicted binary mask a category label from an open vocabulary, we employ text-image discriminative models.
These models~\cite{radford2021clip, jia2021align, mu2021slip}, trained on Internet-scale image-text pairs, have shown strong open-vocabulary classification capabilities. 
They consist of an image encoder $\mathcal{V}$ and a text encoder $\mathcal{T}$. 
Following prior work~\cite{ghiasi2021open, li2022language}, while training, we employ two commonly used supervision signals to learn to predict the category label of each predicted mask. 
Next, we describe how we unify these two training approaches in \ourmethod{}.

\paragraph{Category Label Supervision}
Here, we assume that during training we have access to each mask's ground truth category label. Thus, the training procedure is similar to that of traditional closed-vocabulary training. 
Suppose that there are $K_\text{train} = |\mathbf{C}_{\text{train}}|$ categories in the training set. For each mask embedding feature $z_i$, we dub its corresponding known ground truth category as $y_i \in \mathbf{C}_{\text{train}}$.
We encode the names of all the categories in $\mathbf{C}_{\text{train}}$ with the frozen text encoder $\mathcal{T}$,
and define the set of embeddings of all the training categories' names as
\begin{equation}
\mathcal{T}(\mathbf{C}_\text{train}) \triangleq [\mathcal{T}(c_1), \mathcal{T}(c_2), \dots, \mathcal{T}(c_{K_\text{train}})],
\end{equation} 
where the category name $ c_k \in \mathbf{C}_{\text{train}}$. 
Then we compute the probability of the mask embedding feature $z_i$ belonging to one of the $K_\text{train}$ classes via a classification loss as:
\begin{align}
    & \mathcal{L}_\text{C} = \frac{1}{N}\sum_{i}^{N}\text{CrossEntropy}(\mathbf{p}(z_i, \mathbf{C}_\text{train}), y_i),\\
    \label{eq:prob}
    & \mathbf{p}(z_i, \mathbf{C}_\text{train}) = \text{Softmax}(z_i \cdot \mathcal{T}(\mathbf{C}_\text{train})/\tau),
\end{align}
where $\tau$ is a learnable temperature parameter. 

\paragraph{Image Caption Supervision}
Here, we assume that we do not have any category labels associated with each annotated mask during training. 
Instead, we have access to a natural language caption for each image, and the model learns to classify the predicted mask embedding features using the image caption alone. 
To do so, we extract the nouns from each caption and treat them as the grounding category labels for their corresponding paired image. Following~\cite{gupta2020contrastive, zareian2021open, ghiasi2021open}, we employ a grounding loss to supervise the prediction of the masks' category labels. 
Specifically, given the image-caption pair $(x^{(m)}, s^{(m)})$, suppose that there are $K_\text{word}$ nouns extracted from $s^{(m)}$, denoted as $\mathbf{C}_\text{word} = \{w_k\}_{k=1}^{K_\text{word}}$. 
Suppose further that we sample $B$ image-caption pairs $\{(x^{(m)}, s^{(m)}\}_{m=1}^B$ to form a batch. To compute the grounding loss, we compute the similarity between each image-caption pair as 
\begin{equation}
     g(x^{(m)}, s^{(m)}) = \frac{1}{K}\sum_{k=1}^K\sum_{i=1}^N \mathbf{p}(z_i, \mathbf{C}_\text{word})_k{\cdot} \langle z_i, \mathcal{T}(w_k) \rangle,
\end{equation} 
% where ${\mathbf{p}(z_i, {\mathbf{C_\text{word}}})}$ is defined in Eq.~\ref{eq:prob}. 
where $z_i$ and $\mathcal{T}(w_k)$ are vectors of the same dimension and $\mathbf{p}(z_i, \mathbf{C}_\text{word})_k$ is the $k$-th element of the vector defined in Eq.~\ref{eq:prob} after Softmax.
This similarity function encourages each noun to be grounded by one or a few masked regions of the image and avoids penalizing the regions that are not grounded by any word at all.
Similar to the image-text contrastive loss in~\cite{radford2021clip, jia2021align}, the grounding loss is defined by
\begin{equation}
\begin{split}
    \mathcal{L}_{\text{G}} = &-\frac{1}{B}\sum_{m=1}^B\log \frac{\exp(g(x^{(m)}, s^{(m)}) /\tau) }{\sum_{n=1}^B \exp(g(x^{(m)}, s^{(n)}) /\tau)}\\
        &-\frac{1}{B}\sum_{m=1}^B\log \frac{\exp(g(x^{(m)}, s^{(m)}) /\tau) }{\sum_{n=1}^B \exp(g(x^{(n)}, s^{(m)}) /\tau)},
\end{split}
\end{equation}
where $\tau$ is a learnable temperature parameter.
Finally, note that we train the entire \ourmethod{} model with either $\mathcal{L}_\text{C}$ or $\mathcal{L}_{\text{G}}$, 
together with the class-agnostic binary mask loss. In our experiments, we explicitly state which of these two supervision signals (label or caption) we use for training \ourmethod{} when comparing to the relevant prior works.

% \begin{figure}[t]
%     \centering
%     \includegraphics[width=0.95\linewidth]{figs/inference.pdf}
%     \vspace{-.5em}
%     \caption{
%         \textbf{Open-Vocabulary Inference Pipeline}.  
%         To classify each mask embedding into the testing categories $\mathbf{C}_\text{test}$, we compute its similarity to the text encoder $\mathcal{T}$'s embeddings of category names.
%         % The classification of mask embedding is calculated by the similarity with the text embeddings of category names.
%         Besides the mask embeddings from the text-to-image diffusion model $\{z_i\}_{i=1}^N$, we also perform masked pooling of the features of an image encoder $\mathcal{V}$ from a text-image discriminative model to get $\{z'_i\}_{i=1}^N$.
%         We fuse the predictions of the diffusion model (\textcolor[HTML]{005493}{blue solid path}) and the discriminative model (\textcolor[HTML]{7E7E7E}{grey dash path}) with a geometric mean. 
%     }
%     \vspace{-1.5em}
%     \label{fig:inference}
% \end{figure}

\subsection{Open-Vocabulary Inference}
\label{sec:inference}

During inference (Fig.~\ref{fig:inference}), the set of names of the test categories $\mathbf{C}_\text{test}$ is available, The test categories may be different from the training ones.
Additionally, no caption/labels are available for a test image. 
Hence we pass it through the implicit captioner to obtain its implicit caption; input the two into the diffusion model to obtain the UNet's features; and use the mask generator to predict all possible binary masks of semantic categories in the image.
To classify each predicted mask $m_i$ into one of the test categories, we compute $\mathbf{p}(z_i, \mathbf{C}_\text{test})$ defined in Eq.~\ref{eq:prob} using \ourmethod{} and finally predict the category with the maximum probability. 

In our experiments, we found that the internal representation of the diffusion model is spatially well-differentiated to produce many plausible masks for objects instances. However, its object classification ability can be further enhanced by combining it once again with a text-image discriminative model, e.g., CLIP~\cite{radford2021clip}, especially for open-vocabularies.
To this end, here we leverage a text-image discriminative model's image encoder $\mathcal{V}$ to further classify each predicted masked region of the original input image into one of the test categories. 
Specifically, as Fig.~\ref{fig:inference} illustrates, given an input image $x$, we first encode it into a feature map with the image encoder $\mathcal{V}$ of a text-image discriminative model. 
Then for a mask $m_i$, predicted by \ourmethod{} for image $x$, we pool all the features at the output of the image encoder $\mathcal{V}(x)$ that fall inside the predicted mask $m_i$ to compute a mask pooled image feature for it 
\begin{equation}
z'_i = \text{MaskPooling}(\mathcal{V}(x), m_i).
\end{equation}
We use $\mathbf{p}(z'^M_i, \mathbf{C}_\text{test})$ from Eq.\ref{eq:prob} to compute the final classification probabilities from the text-image discriminative model. 
Finally, we take the geometric mean of the category predictions from the diffusion and discriminative models as the final classification prediction,
\vspace{-1em}
\begin{equation}
   {\mathbf{p}_\text{final} (z_i, \mathbf{C}_\text{test})} \propto {\mathbf{p}(z_i, \mathbf{C}_\text{test})}^\lambda{\mathbf{p}(z'_i, \mathbf{C}_\text{test})}^{(1-\lambda)},
   \vspace{-.5em}
\end{equation} where $\lambda \in [0, 1]$ is a fixed balancing factor.
We find that pooling the masked features is more efficient and yet as effective as the alternative approach proposed in~\cite{ding2022zegformer,gu2021vild}, which crops each of the $N$ predicted masked region's bounding box from the original image and encodes it separately with the image encoder $\mathcal{V}$ (see details in the supplement).

\section{Experiments}

We first introduce our implementation details. Then we compare our results against the state of the art on open-vocabulary panoptic and semantic segmentation.
Lastly, we present ablation studies to demonstrate the effectiveness of the components of our method.

\subsection{Implementation Details}

\paragraph{Architecture}
We use the stable diffusion~\cite{rombach2022ldm} model pre-trained on a subset of the LAION~\cite{schuhmann2021laion} dataset as our text-to-image diffusion model. 
We extract feature maps from every three of its UNet blocks and, like FPN~\cite{lin2017fpn}, resize them to create a feature pyramid.
We set the time step used for the diffusion process to $t=0$, by default. 
We use CLIP~\cite{radford2021clip} as our text-image discriminative model and its corresponding image $\mathcal{V}$ and text $\mathcal{T}$ encoders everywhere. 
We choose Mask2Former~\cite{cheng2022mask2former} as the architecture of our mask generator, and generate $N=100$ binary mask predictions.

\paragraph{Training Details}
We train \ourmethod{} for 90k iterations with images of size $1024^2$ and use large scale jittering~\cite{ghiasi2021simple}.
Our batch size is 64. 
For caption-supervised training, we set $K_\text{word} = 8$.
We use the AdamW~\cite{loshchilov2017decoupled} optimizer with a learning rate $0.0001$ and a weight decay of $0.05$. 
We use the COCO dataset~\cite{lin2014coco} as our training set. 
We utilize its provided panoptic mask annotations as the supervision signal for the binary mask loss.
For training with image captions, for each image we randomly select one caption from the COCO dataset's caption~\cite{chen2015cococap} annotations. 

\paragraph{Inference and Evaluation} 
We evaluate \ourmethod{} on ADE20K~\cite{zhou2019ade} for open-vocabulary panoptic, instance and semantic segmentation; and the Pascal datasets~\cite{everingham2010pascal, mottaghi2014ctx} for semantic segmentation. We also provide the results \ourmethod{} for open-vocabulary object detection and open-world instance segmentation in the supplement. We use only a single checkpoint of \ourmethod{} for mask prediction on all tasks on all datasets. For panoptic segmentation, we report the panoptic quality (PQ) ~\cite{kirillov2019panoptic}, mean average precision (mAP) on the ``thing" categories, and the mean intersection over union (mIoU) metrics (additional SQ and RQ metrics are in the supplement). In panoptic segmentation annotations~\cite{kirillov2019panoptic}, the ``thing'' classes are countable objects like people, animals, \textit{etc.} and the ``stuff'' classes are amorphous regions like sky, grass, \textit{etc.}
Since we train \ourmethod{} with panoptic mask annotations, we can directly infer both instance and semantic segmentation labels with it. 
When evaluating for panoptic segmentation, we use the panoptic test categories as $\mathbf{C}_\text{test}$, and directly classify each predicted mask into the test category with the highest probability. 
For semantic segmentation, we merge all masks assigned to the same ``thing" category into a single one and output it as the predicted mask.

\paragraph{Speed and Model Size}
\ourmethod{} has 28.1M trainable parameters (only 1.8\% of the full model) and 1,493.8M frozen parameters. It performs inference for an image ($1024^2$) at 1.26 FPS on an NVIDIA V100 GPU and uses 11.9 GB memory. 

\begin{table*}[t]
\vspace{-1em}
\tablestyle{8pt}{1.1}
\begin{tabular}{l|ccc|ccc|ccc}
                            & \multicolumn{3}{c|}{Supervision} & \multicolumn{3}{c|}{ADE20K}                     & \multicolumn{3}{c}{COCO}                      \\
Method                      & label    & mask     & caption   & PQ            & mAP           & mIoU          & PQ            & mAP           & mIoU          \\
\shline
MaskCLIP\cite{ding2022open} & \cmark   & \cmark   &           & 15.1          & 6.0           & 23.7          & -             & -             & -             \\
\textbf{\ourmethod{} (Ours)}   & \cmark   & \cmark   &           & \textbf{22.6}          & \textbf{14.4} & \textbf{29.9} & \textbf{55.4} & \textbf{46.0} & \textbf{65.2} \\
\hline
\hline
\textbf{\ourmethod{} (Ours)}   &          & \cmark   & \cmark    & \textbf{23.4} & \textbf{13.9} & \textbf{28.7} & \textbf{45.6} & \textbf{38.4} & \textbf{52.4}
\end{tabular}
\vspace{-1em}
\caption{
    \label{tab:panoptic}
    \textbf{Open-vocabulary panoptic segmentation performance.}
}
\vspace{-.5em}
\end{table*}

\begin{table*}[]
\vspace{-.5em}
\tablestyle{8pt}{1.1}
\begin{tabular}{l|l|ccc|cccccc}
                                    & Training   & \multicolumn{3}{c|}{Supervision} & \multicolumn{6}{c}{mIoU}                                                                      \\
Method                            & Dataset    & label    & mask     & caption   & A-847         & PC-459        & A-150         & PC-59         & PAS-21        & COCO          \\
\shline
SPNet\cite{xian2019spnet}         & Pascal VOC & \cmark   & \cmark   &           & -             & -             & -             & 24.3          & 18.3          & -             \\
ZS3Net\cite{bucher2019zs3net}     & Pascal VOC & \cmark   & \cmark   &           & -             & -             & -             & 19.4          & 38.3          & -             \\
LSeg\cite{li2022language}         & Pascal VOC & \cmark   & \cmark   &           & -             & -             & -             & -             & 47.4          & -             \\
SimBaseline\cite{xu2021simple}    & COCO       & \cmark   & \cmark   &           & -             & -             & 15.3          & -             & 74.5          & -             \\
ZegFormer\cite{ding2022zegformer} & COCO       & \cmark   & \cmark   &           & -             & -             & 16.4          & -             & 73.3          & -             \\
LSeg+\cite{ghiasi2021open}        & COCO       & \cmark   & \cmark   &           & 3.8           & 7.8           & 18.0          & 46.5          & -             & 55.1          \\
MaskCLIP\cite{ding2022open}       & COCO       & \cmark   & \cmark   &           & 8.2           & 10.0          & 23.7          & 45.9          & -             & -             \\
\textbf{\ourmethod{} (Ours)}         & COCO       & \cmark   & \cmark   &           & \textbf{11.1} & \textbf{14.5} & \textbf{29.9} & \textbf{57.3} & \textbf{84.6} & \textbf{65.2} \\
\hline
\hline
GroupViT\cite{xu2022groupvit}     & GCC+YFCC   &          &          & \cmark    & 4.3           & 4.9           & 10.6          & 25.9          & 50.7          & 21.1          \\
OpenSeg\cite{ghiasi2021open}      & COCO       &          & \cmark   & \cmark    & 6.3           & 9.0           & 21.1          & 42.1          & -             & 36.1          \\
\textbf{\ourmethod{} (Ours)}         & COCO       &          & \cmark   & \cmark    & \textbf{11.0} & \textbf{13.8} & \textbf{28.7} & \textbf{55.3} & \textbf{82.7} & \textbf{52.4}
\end{tabular}
\vspace{-1em}
\caption{
    \label{tab:semantic}
    \textbf{Open-vocabulary semantic segmentation performance.}
}
\vspace{-1.2em}
\end{table*}

\begin{figure*}[t]
    \centering
    \vspace{-.5em}
    \includegraphics[width=\linewidth]{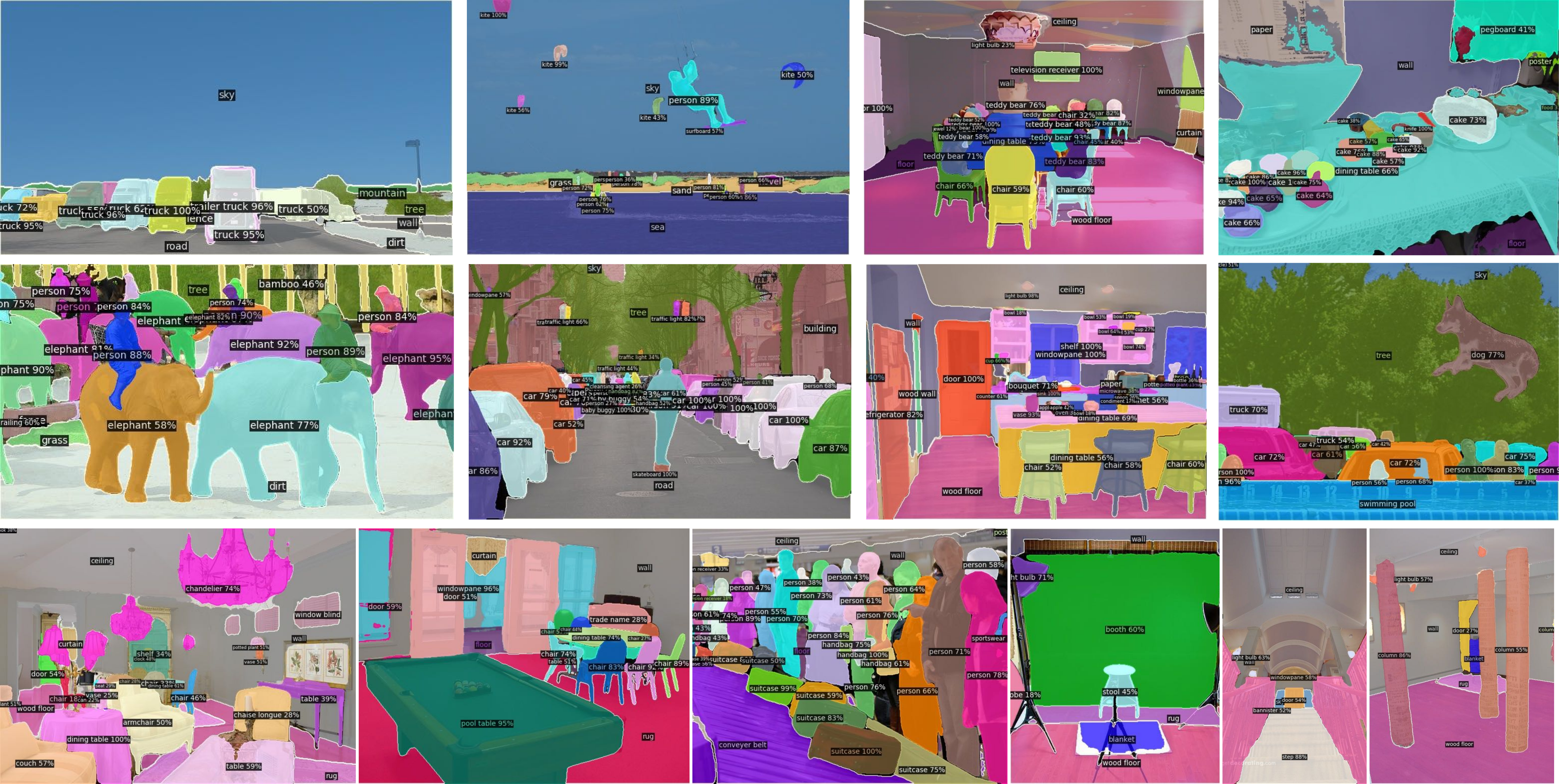}
    \vspace{-1em}
    \caption{
        \textbf{Qualitative Visualization} on COCO (first 2 rows) and ADE20K (last row) validation and test sets. 
        To demonstrate open-vocabulary recognition capability, we merge category names of LVIS, COCO and ADE20K together and perform open-vocabulary inference with ${\sim} 1.5k$ classes directly.
        ``Bamboo'', ``swimming pool'', ``conveyer belt", ``chandelier", ``booth'', ``stool'', ``column'', ``pool table'', ``bannister'', \textit{etc.}, are novel categories from LVIS/ADE20K that are not annotated in COCO. 
        \ourmethod{} shows plausible open-vocabulary panoptic results. The supplement contains more visual results.
    }
    \vspace{-1.2em}
    \label{fig:vis}
\end{figure*}

\subsection{Comparison with State of the Art}
\paragraph{Open-Vocabulary Panoptic Segmentation}
For open-vocabulary panoptic segmentation, we train \ourmethod{} on COCO~\cite{lin2014coco} and test on ADE20K~\cite{zhou2019ade}.
We report results in Table~\ref{tab:panoptic}.
\ourmethod{} outperforms the concurrent work MaskCLIP~\cite{ding2022open} by 8.3 PQ on ADE20K. 
Besides the PQ metric, our approach also surpasses MaskCLIP~\cite{ding2022open} at open-vocabulary instance segmentation on ADE20K, with 8.4 gains in the mAP metric.
The qualitative results can be found in Fig.~\ref{fig:vis} and more in the supplement.

\paragraph{Open-Vocabulary Semantic Segmentation}
We show a comparison of \ourmethod{} to previous work on open-vocabulary semantic segmentation in Table~\ref{tab:semantic}. 
Following the experiment in~\cite{ghiasi2021open}, we evaluate mIoU on 5 semantic segmentation datasets: (a) A-150 with 150 common classes and (b) A-847 with all the 847 classes of ADE20K~\cite{zhou2019ade}, (c) PC-59 with 59 common classes and (d) PC-459 with full 459 classes of Pascal Context~\cite{mottaghi2014ctx}, and (e) the classic Pascal VOC dataset~\cite{everingham2010pascal} with 20 foreground classes and 1 background class (PAS-21).
For a fair comparison to prior work, we train \ourmethod{} with either category or image caption labels. 
\ourmethod{} outperforms the existing state-of-the-art methods on open-vocabulary semantic segmentation~\cite{ghiasi2021open, ding2022open} by a large margin: by 7.6 mIoU on A-150, 4.7 mIoU on A-847, 4.8 mIoU on PC-459 with caption supervision; and by 6.2 mIoU on A-150, 4.5 mIoU on PC-459 with category label supervision, versus the next best method.
Notably, it achieves this despite using supervision from panoptic mask annotations, which is noted to be suboptimal for semantic segmentation~\cite{cheng2022mask2former}. 

We provide comparisons to the state of the art for additional open-vocabulary tasks of object detection and discovery in the supplement.

\subsection{Ablation Study}

To demonstrate the contribution of each component of our method, we conduct an extensive ablation study. 
For faster experimentation, we train \ourmethod{} with $512^2$ resolution images and use image caption supervision everywhere. 

\begin{table}[]
\vspace{-.5em}
\tablestyle{2.2pt}{1.1}
\begin{tabular}{l|c|ccc|ccc}
                                    & Training & \multicolumn{3}{c|}{ADE20K}                    & \multicolumn{3}{c}{COCO}                      \\
Model                              & Data  & PQ            & mAP           & mIoU          & PQ            & mAP           & mIoU          \\
\shline
\multicolumn{8}{l}{Pre-trained with class labels}                                                                                           \\
\hline
DeiT-v3(H)\cite{touvron2022deitv3} & IN-21k   & 21.4          & 11.4          & 28.0          & 41.4          & 29.2          & 52.3          \\
Swin(H)\cite{liu2021swin}          & IN-21k   & 20.9          & 10.7          & 27.7          & 42.4          & 31.6          & 54.0          \\
ConvNeXt(H)\cite{liu2022convnext}  & IN-21k   & 21.0          & 11.0          & 27.8          & 43.1          & 33.1          & 54.3          \\
MViT(H)\cite{li2022mvitv2}         & IN-21k   & 21.1          & 11.6          & 28.1          & 44.0          & 36.3          & \textbf{54.5}          \\
LDM\cite{rombach2022ldm}           & IN-1k    & 20.7          & 10.9          & 26.5          & 41.7          & 35.3          & 50.6          \\
\shline
\multicolumn{8}{l}{Pre-trained with self-supervision}                                                                                                  \\
\hline
MoCo-v3(H)\cite{chen2021mocov3}    & IN-1k    & 19.3          & 9.6           & 25.8          & 37.1          & 26.8          & 47.1          \\
DINO(B)\cite{caron2021dino}        & IN-1k    & 20.6          & 10.5          & 26.3          & 39.5          & 29.8          & 49.5          \\
MAE(H)\cite{he2022mae}             & IN-1k    & 21.5          & 10.9          & 27.6          & 37.9          & 31.6          & 46.3          \\
BEiT-v2(H)\cite{peng2022beitv2}    & IN-21k   & 21.4          & 11.4          & 28.0          & 41.4          & 29.2          & 52.3          \\
\shline
\multicolumn{8}{l}{Pre-trained with text}                                                                                                  \\
\hline
CLIP(L)\cite{radford2021clip}      & WIT      & 20.4          & 9.6           & 27.0          & 40.6          & 26.7          & 52.1          \\
CLIP(H)\cite{radford2021clip}      & LAION    & 21.2          & 10.8          & 28.1          & 41.0          & 27.9          & 52.1          \\
\textbf{\ourmethod{}}                & LAION    & \textbf{23.3} & \textbf{13.0} & \textbf{29.2} & \textbf{44.2} & \textbf{38.3} & 53.8
\end{tabular}
\vspace{-1em}
\caption{
    \label{tab:pretrain}
    \textbf{Comparison with the state-of-the-art visual representations.} B, L, H in the parentheses denote the model's size. 
}
\vspace{-2em}
\end{table}

\paragraph{Visual Representations} 
We compare the internal representation of text-to-image diffusion models to those of other state-of-the-art pre-trained discriminative and generative models. We evaluate various discriminative models trained with full label, text or self-supervision.
In all experiments we freeze the weights of the pre-trained models and use exactly the same training hyperparameters and mask generator as in our method. 
For each supervision category we select the best-performing and largest publicly available discriminative models.  
We observe from Table~\ref{tab:pretrain} that \ourmethod{} outperforms all other models in terms of PQ on both datasets. 
To offset any potential bias arising from the larger size of the LAION dataset (2B image-caption pairs) with which the stable diffusion model is trained, versus the smaller datasets used to train the discriminative models, we also compare to CLIP(H)~\cite{ilharco2021openclip,radford2021clip}. It is trained on an equal-sized LAION~\cite{schuhmann2021laion} dataset. Despite both models being trained on the same data, our diffusion-based method outperforms CLIP(H) by a large margin on all metrics. This demonstrates that the diffusion model's internal representation is indeed superior for open-vocabulary segmentation that that of discriminative pre-trained models.

The recent DDPMSeg\cite{baranchuk2021ddpmseg} model  is somewhat related to our model. Besides us, it is the only prior work that uses diffusion models and obtains state-of-the-art performance on label-efficient segmentation learning. 
Since DDPMSeg relies on category specific diffusion models it is not designed for open-world panoptic segmentation. Hence its direct comparison against our approach is not feasible.
As an alternative, we compare against the internal representations of a class-conditioned generative model~\cite{rombach2022ldm} trained on more categories from ImageNet\cite{deng2009imagenet} (LDM row in  Table~\ref{tab:pretrain}). 
Not surprisingly, we find that despite both generative models being diffusion-based, our approach of using a model trained on Internet-scale data is more effective at generalizing to open-vocabulary categories.

\begin{table}[]
\vspace{-.5em}
\tablestyle{4pt}{1.1}
\begin{tabular}{ll|ccc|ccc}
    &                                  & \multicolumn{3}{c|}{ADE20K}                    & \multicolumn{3}{c}{COCO}                      \\
    & Captioner                        & PQ            & mAP           & mIoU          & PQ            & mAP           & mIoU          \\
\shline
(a) & Empty                            & 21.8          & 11.8          & 27.3          & 43.5          & 37.0          & 52.3          \\
(b) & Heuristic\cite{zeng2022socratic} & 22.2          & 12.1          & 28.1          & 44.0          & 36.3          & 53.3          \\
(c) & BLIP\cite{li2022blip}            & 22.3          & 12.4          & 28.2          & 44.1          & 37.1          & 53.6          \\
(d) & Implict                          & \textbf{23.3} & \textbf{13.0} & \textbf{29.2} & \textbf{44.2} & \textbf{38.3} & \textbf{53.8}
\end{tabular}
\vspace{-1em}
\caption{
    \label{tab:captioner}
    \textbf{Ablation results of different caption generators.}
}
\vspace{-2em}
\end{table}
\paragraph{Captioning Generators}
As discussed in Sec.~\ref{sec:diffusion}, the internal features of a text-to-image diffusion model are dependent on the embedding of the input caption. 
To derive the optimal set of features for our downstream task, we introduce a novel implicit captioning module to directly generate implicit text embeddings from an image. 
This module also facilitates inference on images sans paired captions at test time.
Here, we construct several baselines to show the effectiveness of our implicit captioning module.
The results are shown in Table~\ref{tab:captioner}. 
The various alternatives that we compare are: providing an empty string to the text encoder for any given image, such that the text embedding for all images is fixed (row (a)); employing two different off-the-shelf image captioning networks to generate an explicit caption for each image on-the-fly (rows (b) and (c)), where (c)\cite{li2022blip} is trained on the COCO caption dataset, while (b)\cite{zeng2022socratic} is not; and our proposed implicit captioning module (row (d)). Overall, we find that using an explicit/implicit caption is better than using empty text. Furthermore, (c) improves over (b) on COCO but has similar PQ on ADE20K. 
It may be because the pre-trained BLIP~\cite{li2022blip} model does not see ADE20K's image distribution during training and hence it cannot output high-quality captions for it. 
Lastly, since our implicit captioning module derives its caption from a text-image discriminative model trained on Internet-scale data, it is able to generalize best among all variants compared.

\begin{table}[]
\tablestyle{6pt}{1.1}
\begin{tabular}{c|ccc|ccc}
          & \multicolumn{3}{c|}{ADE20K}                    & \multicolumn{3}{c}{COCO}                      \\
time step  & PQ            & mAP           & mIoU          & PQ            & mAP           & mIoU          \\
\shline
0         & \textbf{23.3} & \textbf{13.0} & 29.2 & \textbf{44.2} & \textbf{38.3} & \textbf{53.8} \\
100       & 22.8          & 12.5          & 29.3          & 43.2          & 36.4          & 52.3          \\
200       & 21.5          & 11.9          & 28.0          & 42.4          & 35.1          & 51.7          \\
500       & 20.9          & 11.1          & 27.0          & 38.2          & 31.1          & 47.6          \\
0+100+200 & 23.1          & 12.9          & \textbf{29.7}          & 43.7          & 37.4          & 53.0          \\
\hline
learnable & 22.8          & 12.9          & 29.2          & 44.0          & 37.5          & 53.4            
\end{tabular}
\vspace{-.5em}
\caption{
    \label{tab:step}
    \textbf{Ablation results of different diffusion time steps.} 0+100+200 denotes the concatenation of the features at time steps 0, 100, and 200.
}
\vspace{-1.5em}
\end{table}
\paragraph{Diffusion Time Steps}
We also study which diffusion step(s) are most effective for extracting features from, similarly to DDPMSeg~\cite{baranchuk2021ddpmseg}.
The noise process is defined in Eq.\ref{eqn:noise}. The larger the $t$ value is, the larger the noise distortion added to the input image is.
In stable diffusion~\cite{rombach2022ldm} there are a 1000 total time steps. 
From Table~\ref{tab:step}, all metrics decrease as $t$ increases and the best results are for $t{=}0$ (our final value). 
Concatenating 3 time steps, 0, 100, 200, yields a similar accuracy to $t{=}0$ only, but is $3{\times}$ slower. 
We also train our model with $t$ as a learnable parameter, and find that many random training runs all converge to a value close to zero, further validating our optimal choice of $t{=}0$.

\begin{table}[]
\tablestyle{2.5pt}{1.1}
\begin{tabular}{cc|ccc|ccc}
\multicolumn{2}{c|}{model}  & \multicolumn{3}{c|}{ADE20K}                    & \multicolumn{3}{c}{COCO}                      \\
diffusion & discriminative & PQ            & mAP           & mIoU          & PQ            & mAP           & mIoU          \\
\shline
          & \cmark         & 15.0          & 9.6           & 17.5          & 26.5          & 23.5          & 23.6          \\
\cmark    &                & 20.1          & 10.3          & 24.4          & 42.3          & 37.8          & 52.0          \\
\cmark    & \cmark         & \textbf{23.3} & \textbf{13.0} & \textbf{29.2} & \textbf{44.2} & \textbf{38.3} & \textbf{53.8}
\end{tabular}
\vspace{-.5em}
\caption{
    \label{tab:fusion}
    \textbf{Ablation results of fusing class predictions of diffusion and discriminative models.}
}
\vspace{-1.5em}
\end{table}

% \vspace{-0.25em}
\paragraph{Mask Classifiers}
For final open-vocabulary classifciation (Fig.~\ref{fig:inference}), we fuse class prediction from the diffusion and discriminative models.
We report their individual performance in Table~\ref{tab:fusion}. 
Individually the diffusion approach performs better on both datasets than the discriminative only approach. Nevertheless, fusing both together results in higher values on both ADE20K and COCO. Finally, note that even without fusion, our diffusion-only method already surpasses existing methods (see Tables~\ref{tab:panoptic}, \ref{tab:semantic}).

\section{Conclusion}

We take the first step in leveraging the frozen internal representation of large-scale text-to-image diffusion models for downstream recognition tasks.
\ourmethod{} shows the great potential of text-to-image generation models in open-vocabulary segmentation tasks and establishes a new state of the art.
This work demonstrates that text-to-image diffusion models are not only capable of generating plausible image but also of learning rich semantic representations.
It opens up a new direction for how to effectively leverage the internal representation of text-to-image models for other tasks as well in the future.

{\footnotesize \noindent \textbf{Acknowledgements.} 
We thank Golnaz Ghiasi for providing the prompt engineering labels for evaluation. 
Prof. Xiaolong Wang’s laboratory was supported, in part, by NSF CCF-2112665 (TILOS), NSF CAREER Award IIS-2240014, DARPA LwLL, Amazon Research Award, Adobe Data Science Research Award, and Qualcomm Innovation Fellowship.
}

%%%%%%%%% REFERENCES
{\small
\bibliographystyle{ieee_fullname}
\bibliography{egbib}
}

% supp
% reset Figure number: e.g., Figure B.1.
\renewcommand\thefigure{\thesection.\arabic{figure}}
\renewcommand\thetable{\thesection.\arabic{table}}
\setcounter{figure}{0} 
\setcounter{table}{0} 

\appendix

% In the supplementary material, we first describe additional implementation details of \ourmethod{} and its experimental setting. 
% Then we show more qualitative results of open-vocabulary panoptic segmentation on the COCO\cite{lin2014coco}, ADE20K\cite{zhou2019ade} and Ego4D\cite{grauman2022ego4d} datasets. Lastly, we provide quantitative results of the performance of \ourmethod{} for the additional tasks of open-vocabulary object detection and open-world instance segmentation.

In this supplement we provide additional implementation details; quantitative experimental and qualitative visual results.

\section{Implementation Details}

We open-source our code and models at \href{https://github.com/NVlabs/ODISE}{https://github.com/NVlabs/ODISE}.

\paragraph{Training}
We train \ourmethod{} for 90k iterations with images of size $1024^2$ and use large scale jittering~\cite{ghiasi2021simple} with random scales between $[0.1{-}2.0]$ as data augmentation.
We use 32 NVIDIA V100 GPUs with 2 images per GPU with an effective batch size is 64. 
We use the AdamW~\cite{loshchilov2017decoupled} optimizer with a learning rate $0.0001$ and a weight decay of $0.05$. 
We use a step learning rate schedule and reduce the learning rate by a factor of $10$ at 81k and 86k iterations.
We set the balancing factor between the diffusion and discriminative models to $\lambda=0.65$ for all tasks.
Following \cite{cheng2021maskformer,cheng2022mask2former, carion2020detr}, we use Hungarian matching to match the predicted masks to the ground-truth ones. We compute the training losses between the matched pairs.

\paragraph{Open-Vocabulary Inference}
An object can often be described by more than one possible description, e.g., the dog category could be described by ``dog'' or ``puppy''.
We use the same prompt engineering strategy as in~\cite{ghiasi2021open} to create an ensemble of text prompts for each test category and predict the category with the maximum probability.

\paragraph{Speed and Model Size}
It takes 5.3 days to train \ourmethod{} for 90k iterations on the COCO dataset. It has 28.1M trainable parameters (only 1.8\% of the full model) and 1,493.8M frozen parameters (including Stable Diffusion and CLIP).
It performs single image inference at 1.26 FPS on an NVIDIA V100 GPU and uses 11.9 GB memory with an image of size $1024^2$. %Compared to CLIP(H) from Table~3 of the main paper) with an FPS of 1.33 FPS (the proxy to MaskCLIP), \ourmethod{} only increases computations by 5\%.
We also replace the bounding box cropping proposed in~\cite{gu2021vild} that runs at 0.38 FPS, with mask feature pooling described in Section~3.6 of the main paper. Mask pooling yields a 3x speedup, while maintaining similar PQ on ADE20K: 23.4 for mask pooling versus 23.7 for bounding box cropping.

\section{Experiments}

\subsection{Comparison with State of the Art}
\paragraph{Open-Vocabulary Panoptic Segmentation}

Besides panoptic quality (PQ), we additionally report the detailed metrics of segmentation quality (SQ) and recognition quality (RQ) for \ourmethod{} and MaskCLIP~\cite{ding2022open} on both the thing (${\text{Th}}$) and stuff (${\text{St}}$) categories of the ADE20K dataset in Table~\ref{tab:ade_breakdown}. Here, all models were trained on COCO. \ourmethod{} outperforms MaskCLIP~\cite{ding2022open} w.r.t. all metrics.

\begin{table}[!h]
\tablestyle{1.3pt}{1.1}
   % \vspace{-1em}
   \begin{tabular}{l|ccc|ccc|ccc}
   Method                    & PQ   & PQ$^{\text{Th}}$ & PQ$^{\text{St}}$ & SQ   & SQ$^{\text{Th}}$ & SQ$^{\text{St}}$ & RQ   & RQ$^{\text{Th}}$ & RQ$^{\text{St}}$ \\
   \shline
   MaskCLIP                  & 15.1          & 13.5             & 18.3             & 70.5          & 70.0             & 71.4             & 19.2          & 17.5             & 22.7             \\
   \textbf{\ourmethod{} (Ours)} & \textbf{23.4} & \textbf{21.9}    & \textbf{26.6}    & \textbf{78.1} & \textbf{77.7}    & \textbf{78.8}    & \textbf{28.3} & \textbf{26.6}    & \textbf{31.6}   
   \end{tabular}
   \vspace{-.5em}
   \caption{
      \label{tab:ade_breakdown}
      \textbf{Detailed panoptic segmentation metrics on ADE20K.} \ourmethod{} outperforms MaskCLIP~\cite{ding2022open} w.r.t. all metrics.
   }
   \vspace{-1em}
\end{table}

\begin{table}[!h]
\tablestyle{6pt}{1.1}
\vspace{-1.2em}
\begin{tabular}{l|ccc|ccc}
                              & \multicolumn{3}{c|}{Cityscapes}                & \multicolumn{3}{c}{Mapillary Vistas}          \\
   Method                    & PQ            & SQ            & RQ            & PQ            & SQ            & RQ            \\
   \shline
   CLIP(H)                   & 18.5          & 69.4          & 24.2          & 11.7          & 60.5          & 15.1          \\
   \textbf{\ourmethod{} (Ours)} & \textbf{23.9} & \textbf{75.3} & \textbf{29.0} & \textbf{14.2} & \textbf{61.0} & \textbf{17.2}
\end{tabular}
\vspace{-.5em}
\caption{
    \label{tab:cs}
    \textbf{Results of panontic segmentation on Cityscapes and Mapillary Vistas.} \ourmethod{} outperforms CLIP(H) by a large margin on both datasets.
}
\vspace{-1.em}
\end{table}

% \subsection{Results on More Panoptic Segmentation Datasets}
We also evaluate \ourmethod{} trained on COCO on the Cityscapes~\cite{cordts2016cityscapes} and Mapillary Vistas~\cite{neuhold2017mapillary} datasets in Table~\ref{tab:cs}. Since the source code for MaskCLIP~\cite{ding2022open} is not  publicly available, we regard \ourmethod{}'s implementation with CLIP(H) features (from Table~3 of the main paper) as a close proxy to MaskCLIP and compare against it (Table~\ref{tab:cs}). Here too, \ourmethod{}, which is based on diffusion features, outperforms its CLIP(H) variants by large margins. Note that in this experiment, we use the original text labels provided with the respective test datasets and didn't carefully select the category names for computing the text embedding. Hence, the results could be further improved if categories like ``terrain" are converted into more detailed descriptions.

% \subsection{Swap Training and Evaluation Dataset}

Finally, to additionally verify the effectiveness of \ourmethod{}, we also swap the training and evaluation datasets, i.e., we train on ADE20K and evaluated on COCO, and report the results in Table~\ref{tab:swap}. Here too, we regard the variant of \ourmethod{} with CLIP(H) features as a proxy to MaskCLIP~\cite{ding2022open} and compare against it. \ourmethod{} outperforms its CLIP(H) variant by a large margin.

\begin{table}[!h]
\tablestyle{6pt}{1.1}
\vspace{-1em}
\begin{tabular}{l|ccc|ccc}
                              & \multicolumn{3}{c|}{COCO}                & \multicolumn{3}{c}{ADE20K}          \\
   Method                    & PQ            & SQ            & RQ            & PQ            & SQ            & RQ            \\
   \shline
   CLIP(H)                   & 20.7          & 72.6          & 26.5          & 25.7          & 72.3          & 32.1          \\
   \textbf{\ourmethod{} (Ours)} & \textbf{25.0} & \textbf{79.4} & \textbf{30.4} & \textbf{31.4} & \textbf{77.9} & \textbf{36.9}
\end{tabular}
\caption{
    \label{tab:swap}
    \textbf{Results of swapped training on ADE20K and testing on COCO.} \ourmethod{} outperforms CLIP(H) by a large margin on both datasets.
}
\vspace{-1em}
\end{table}

\paragraph{Open-Vocabulary Object Detection}
We also evaluate \ourmethod{} for the task of open-vocabulary object detection on the LVIS~\cite{gupta2019lvis} dataset (Table~\ref{tab:object}). By regarding all categories to belong to ``things", we directly evaluate on LVIS's object detection labels, which contain annotations for 1203 fine-grained categories for COCO~\cite{lin2014coco} images. 
For this task, we measure
$\text{mAP}_r$, which denotes the mAP score on 337 rare categories only. 
We evaluate \ourmethod{} trained with both types of supervision: mask category labels or image captions.
\ourmethod{} outperforms MaskCLIP\cite{ding2022open} by a large margin w.r.t. both mAP and $\text{mAP}_r$.
Note that the validation split of LVIS\cite{gupta2019lvis} has overlapping images with COCO\cite{lin2014coco}'s training split, but the category labels of LVIS are only available during inference.

\begin{table}[h]
\tablestyle{7pt}{1.1}
\begin{tabular}{l|ccc|cc}
                            & \multicolumn{3}{c}{Supervision} & \multicolumn{2}{c}{LVIS}       \\
Method                      & label    & mask     & caption   & mAP           & $\text{mAP}_r$ \\
\shline
MaskCLIP\cite{ding2022open} & \cmark   & \cmark   &           & 8.4           & -              \\
\hline
\hline
\textbf{\ourmethod{} (Ours)}   & \cmark   & \cmark   &           & 15.4          & 19.4           \\
\textbf{\ourmethod{} (Ours)}   &          & \cmark   & \cmark    & \textbf{17.1} & \textbf{21.1} 
\end{tabular}
\vspace{-.5em}
\caption{
\label{tab:object}
  \textbf{Open-Vocabulary Object Detection.} $\text{mAP}_r$ denotes the mAP score for 337 rare categories only. \ourmethod{} surpasses MaskCLIP by a large margin, both with category label and caption  during training.
}
\vspace{-1em}
\end{table}

\paragraph{Open-World Instance Segmentation}
The task of open-world instance segmentation aims at discovering at test time, all plausible instance masks that may be present in an image in a class-agnostic manner. 
We also evaluate \ourmethod{} in for this task.
Following~\cite{wang2022ggn}, we report the average recall of 100 mask proposals (AR@100) on the UVO~\cite{wang2021uvo} and ADE20K~\cite{zhou2019ade} datasets.
As reported in Table~\ref{tab:instance}, here too we outperform the existing state of the art~\cite{wang2022ggn} by 14.3 points on UVO and 9.3 points on ADE20K.
It demonstrates that with the internal representation of pre-trained text-to-image diffusion models it is plausible to discover open-world instances.

\begin{table}[h]
\tablestyle{7pt}{1.1}
\begin{tabular}{l|ccc}
                            & \multicolumn{3}{c}{AR@100}                    \\
Method                    & UVO           & ADE20K        & COCO          \\
\shline
LDET\cite{kim2022ldet}    & 42.6          & -             & -             \\
GGN\cite{wang2022ggn}     & 43.4          & 21.0          & -             \\
\textbf{\ourmethod{} (Ours)} & \textbf{57.7} & \textbf{30.3} & \textbf{56.6}
\end{tabular}
\vspace{-.5em}
\caption{
    \label{tab:instance}
    \textbf{Open-world Instance Segmentation.} \ourmethod{} outperforms GGN on discovering open-world instances on both the UVO and ADE20K datasets.
}
\vspace{-1em}
\end{table}

\subsection{Ablation Study}

\begin{figure}[!h]
   \centering
   \includegraphics[width=.95\linewidth, trim={0 0 0 3em}, clip]{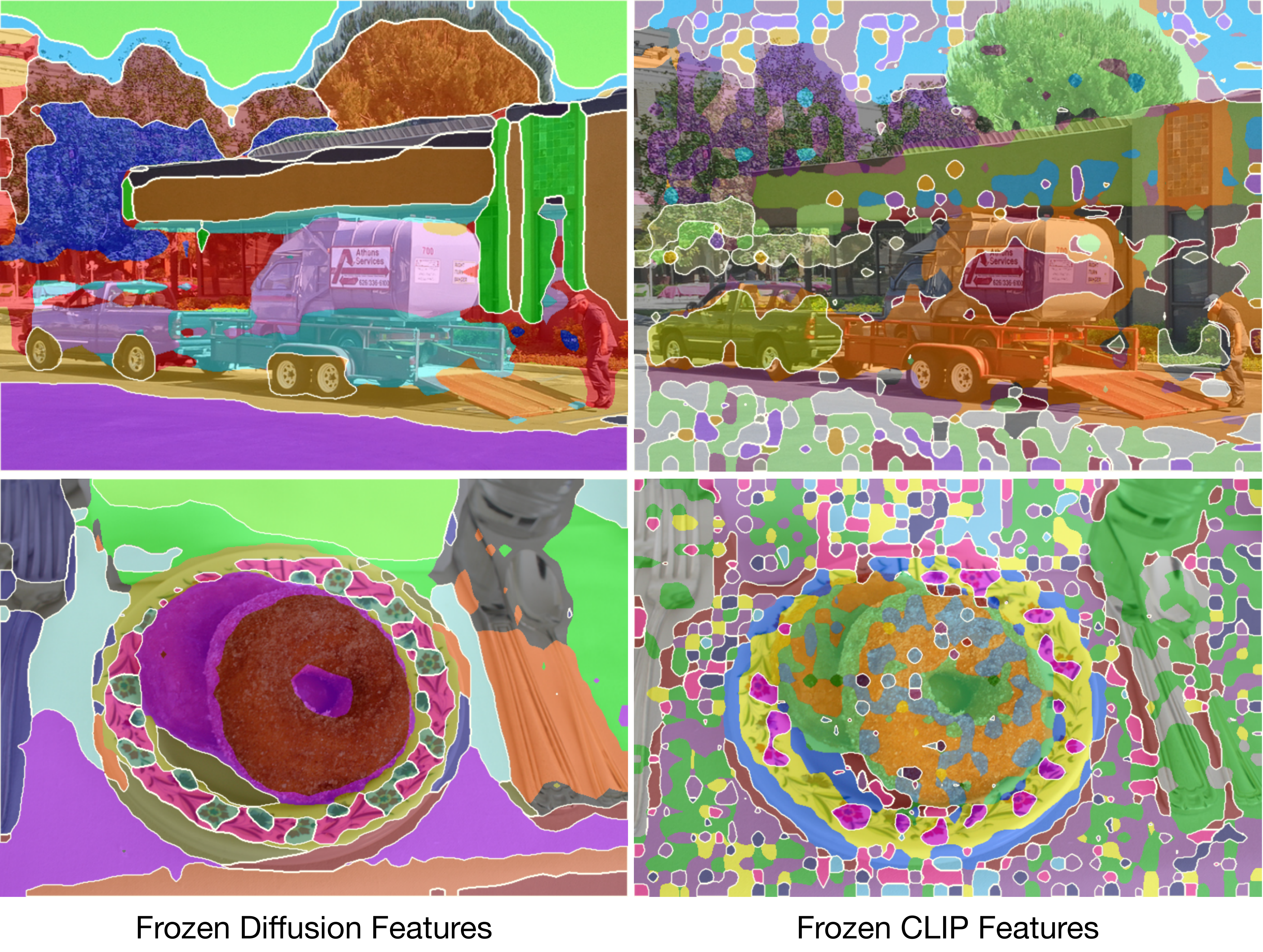}
   \vspace{-.5em}
   \caption{
      \label{fig:cluster_compare}
      \textbf{K-mean clustering of text-to-image diffusion and CLIP models' internal features}. The internal features of the diffusion model are much more semantically differentiated than those from CLIP.
   }
%    \vspace{-.5em}
\end{figure}

\paragraph{Visual Representations}
In Fig.~\ref{fig:cluster_compare} we show k-means clustering of the text-to-image diffusion model's and CLIP's frozen internal features; diffusion features are much more semantically differentiated. Quantitative comparisons of \ourmethod{} and its CLIP(H) variant in Table~3 of the main paper and Table~\ref{tab:cs}  and Table~\ref{tab:swap} further substantiate diffusion features' superiority over those of CLIP's. 

\begin{figure*}[t]
    \centering
    \includegraphics[width=0.98\linewidth]{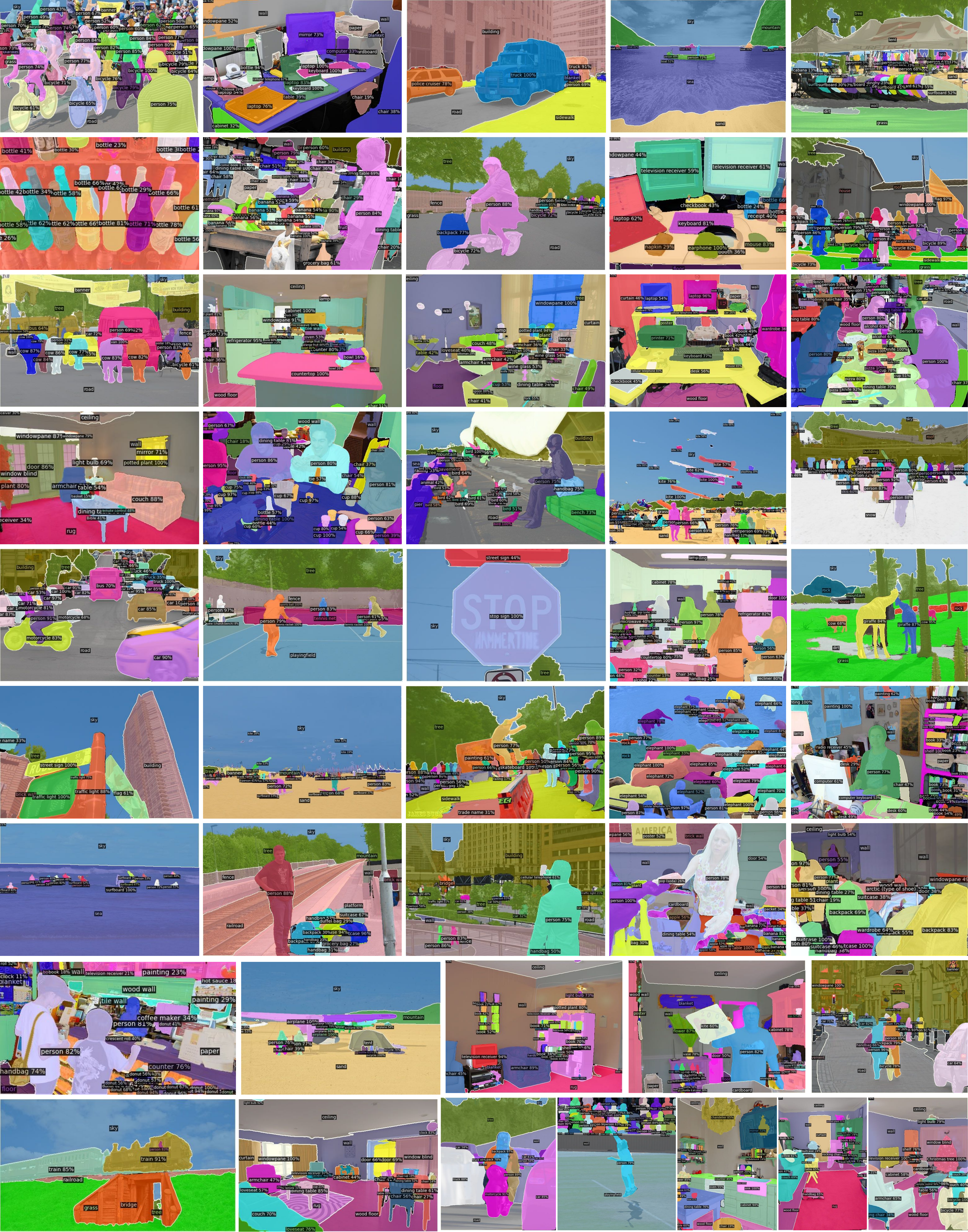}
    \caption{
        \textbf{Qualitative visualization of open-vocabulary panoptic segmentation results on COCO}. 
    }
    \vspace{-1.2em}
    \label{fig:vis_coco_supp}
\end{figure*}

\begin{figure*}[t]
    \centering
    % \vspace{-.5em}
    \includegraphics[width=0.98\linewidth]{figs/vis_ade_supp.pdf}
    % \vspace{-1em}
    \caption{
        \textbf{Qualitative visualization of open-vocabulary panoptic segmentation results on ADE20K}. 
    }
    \vspace{-1.2em}
    \label{fig:vis_ade_supp}
\end{figure*}

\begin{figure*}[t]
    \centering
    % \vspace{-.5em}
    \includegraphics[width=0.98\linewidth]{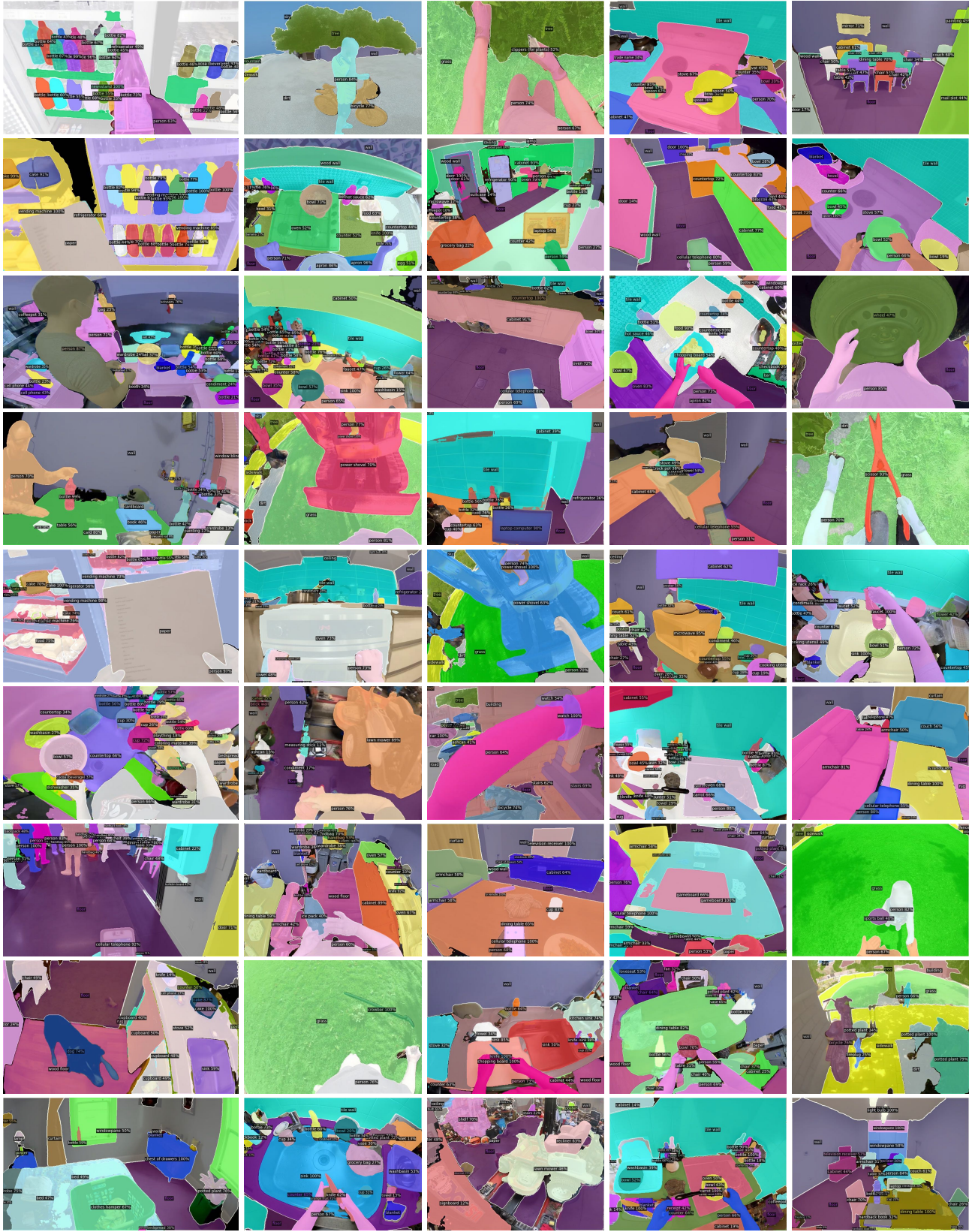}
    % \vspace{-1em}
    \caption{
        \textbf{Qualitative visualization of open-vocabulary panoptic segmentation results on Ego4D}. 
    }
    \vspace{-1.2em}
    \label{fig:vis_ego4d_supp}
\end{figure*}
\section{Qualitative Results}

To demonstrate the open-vocabulary recognition capabilities of \ourmethod{}, we merge the category names from LVIS~\cite{gupta2019lvis}, COCO~\cite{lin2014coco}, ADE20K~\cite{zhou2019ade} together and perform open-vocabulary inference with ${\sim} 1.5k$ test classes.
We only train \ourmethod{} on COCO's~\cite{lin2014coco} training dataset and evaluate open-vocabulary panoptic inference on ADE20K~\cite{zhou2019ade} and Ego4D\cite{grauman2022ego4d}.
The qualitative results on COCO's~\cite{lin2014coco} validation dataset, ADE20K~\cite{zhou2019ade} and Ego4D~\cite{grauman2022ego4d} are shown in Fig.~\ref{fig:vis_coco_supp}, Fig.~\ref{fig:vis_ade_supp} and Fig.~\ref{fig:vis_ego4d_supp}, respectively. 
Most categories, e.g., ``police cruiser'', ``flag'', ``conveyor belt", ``chandelier", ``aquarium", ``grocery bag'', ``power shovel'', \textit{etc.}, are novel categories from LVIS~\cite{gupta2019lvis} or ADE20K~\cite{zhou2019ade} that are not annotated in COCO~\cite{lin2014coco}. 
It is worth noting that Ego4D\cite{grauman2022ego4d} is a video dataset, which consists of diverse ego-centric videos. 
Despite the large domain gap between the testing dataset Ego4D~\cite{grauman2022ego4d} and our training dataset COCO~\cite{lin2014coco}, \ourmethod{} still outputs good-quality plausible panoptic segmentation results on Ego4D's novel categories.

\section{Limitations and Future Work}
In the current datasets, the category definitions are sometimes ambiguous and non-exclusive, e.g., in ADE20K, ``tower'' is often mis-classified as ``building''.
Although this could be mitigated by prompt and ensemble engineering, how category definitions affect evaluation accuracy, would be interesting to analyze in the future.

\section{Ethics Concerns}
The text-to-image diffusion model that we use is pre-trained with web-crawled image-text pairs collected by previous works. Despite applying filtering, there may still be potential bias in its internal representation.

\end{document}